\pdfoutput=1

\documentclass[11pt]{article}

\usepackage{amsmath,amsfonts,bm}

\def\eqref#1{equation~\ref{#1}}

\def\1{\bm{1}}

\DeclareMathAlphabet{\mathsfit}{\encodingdefault}{\sfdefault}{m}{sl}
\SetMathAlphabet{\mathsfit}{bold}{\encodingdefault}{\sfdefault}{bx}{n}

\usepackage{color}
\usepackage{times}
\usepackage{epsfig}
\usepackage{graphicx}
\usepackage{amsmath}
\usepackage{amsthm}
\usepackage{amssymb}
\usepackage{mathtools}
\usepackage{multirow}
\usepackage{bbm}
\usepackage{dsfont}

\usepackage[table, dvipsnames]{xcolor}

\usepackage[utf8]{inputenc}
\usepackage{booktabs,siunitx}
\usepackage{multirow}
\usepackage{multicol}
\usepackage{amsmath}
\usepackage{subcaption}
\usepackage{caption}
\usepackage{graphicx}
\usepackage{pifont}

\usepackage{tikz}
\usetikzlibrary{shapes,arrows,patterns}
\usetikzlibrary{positioning}
\usetikzlibrary{calc}

\newcommand{\cut}[1]{}

\newcommand{\postspace}{\vskip -3mm}
\newcommand{\minipostspace}{\vskip -2mm}

\definecolor{red}{RGB}{255, 117, 115}
\definecolor{green}{RGB}{171, 255, 175}

\usepackage[preprint]{acl}

\usepackage{times}
\usepackage{latexsym}

\usepackage[T1]{fontenc}

\usepackage[utf8]{inputenc}

\usepackage{microtype}

\usepackage{inconsolata}

\usepackage{graphicx}
\definecolor{cadmiumgreen}{rgb}{0.0, 0.42, 0.24}
\definecolor{cardinal}{rgb}{0.77, 0.12, 0.23}
\definecolor{cadmiumred}{rgb}{0.89, 0.0, 0.13}

\usepackage{tcolorbox}
\newtcolorbox[list inside=prompt,auto counter,number within=section]{prompt}[1][]{
    fontupper=\ttfamily\footnotesize,
    boxsep=5pt,
    left=0pt,
    right=0pt,
    top=0pt,
    bottom=0pt,
    boxrule=1pt,
    #1,
}
\title{Multi-Conditional Ranking with Large Language Models}

\author{Pouya Pezeshkpour\\
Megagon Labs\\
\texttt{pouya@megagon.ai} \\
\And
Estevam Hruschka\\
Megagon Labs \\
\texttt{estevam@megagon.ai} \\
}

\begin{document}
\maketitle
\begin{abstract}
Utilizing large language models (LLMs) to rank a set of items has become a common approach in recommendation and retrieval systems. Typically, these systems focus on ordering a substantial number of documents in a monotonic order based on a given query. However, real-world scenarios often present a different challenge: ranking a comparatively smaller set of items, but according to a variety of diverse and occasionally conflicting conditions. 
In this paper, we define and explore the task of multi-conditional ranking by introducing MCRank, a benchmark tailored for assessing multi-conditional ranking across various item types and conditions. Our analysis of LLMs using MCRank indicates a significant 
decrease in performance as the number and complexity of items and conditions grow. To overcome this limitation, we propose a novel decomposed reasoning method, 
consisting of \textbf{EX}tracting and \textbf{S}orting the conditions, and then \textbf{I}teratively \textbf{R}anking the items (EXSIR).
Our extensive experiments show that this decomposed reasoning method enhances LLMs' performance significantly, achieving up to a 14.4\% 
improvement over existing LLMs. We also provide a detailed analysis of LLMs performance across various condition categories, and examine the effectiveness of decomposition step. Furthermore, we compare our method with existing approaches such as Chain-of-Thought and existing ranking models, demonstrating the superiority of our approach and complexity of MCR task. We released our dataset and code\footnote{\url{https://github.com/megagonlabs/MCR}}.
\end{abstract}

\section{Introduction}
The rapid advancement of autoregressive Large Language Models (LLMs) has significantly enhanced our ability to understand and solve NLP related tasks \citep{chowdhery2022palm,touvron2023llama,openai2023gpt-4,team2023gemini}. Among these tasks, document ranking plays a crucial role in recommendation and retrieval systems \citep{wu2023survey,zhu2023large}. While there has been a considerable advancement in ranking extensive document collections given a query \citep{khattab2020colbert,zhuang2023setwise,qin2023large}, the nuanced task of ranking a smaller set of items based on multiple conditions---a critical requirement in numerous real-world applications---has not been addressed in prior research.

Ranking a set of items according to multiple conditions has vast implications across various fields and applications. In recommendation systems, for instance, once the top candidates are shortlisted, the user experience can be significantly enhanced by offering the capability to re-rank these candidates based on specific conditions, such as genres and categories. 
In the realm of education, this task can be applied to the ranking of questions, enabling educators to prioritize and arrange questions effectively according to different criteria, such as subject matter. Moreover, in competitive job markets, multi-conditional ranking is crucial for matching resumes to job postings, prioritizing factors like skills and experience.

\begin{figure*}[t!]
    \centering
    \begin{subfigure}[b]{0.24\textwidth}
        \centering
        \includegraphics[width=\linewidth]{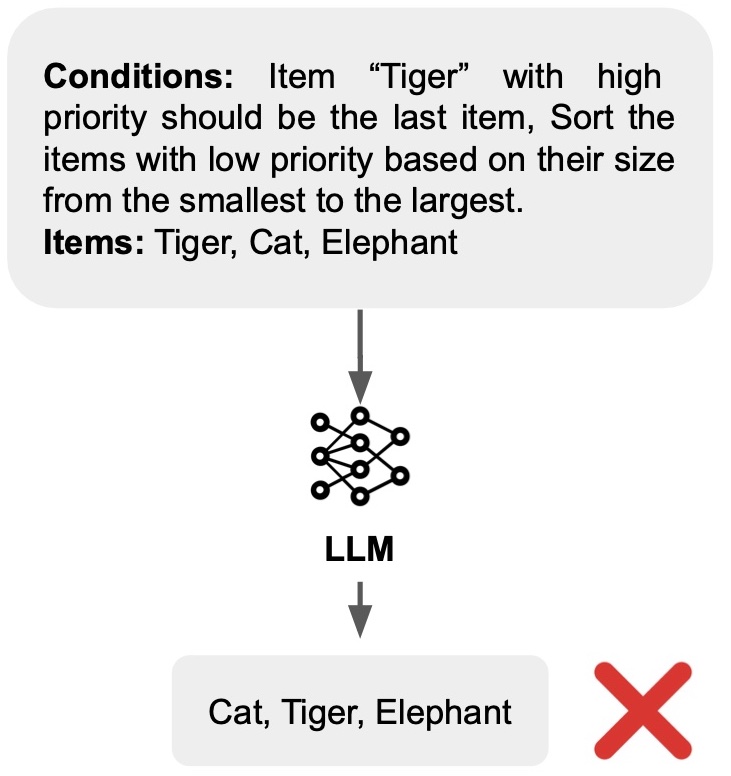}  
        \caption{Base Approach}
    \end{subfigure}%
    ~ 
    \begin{subfigure}[b]{0.65\textwidth}
        \centering
        \includegraphics[width=\linewidth]{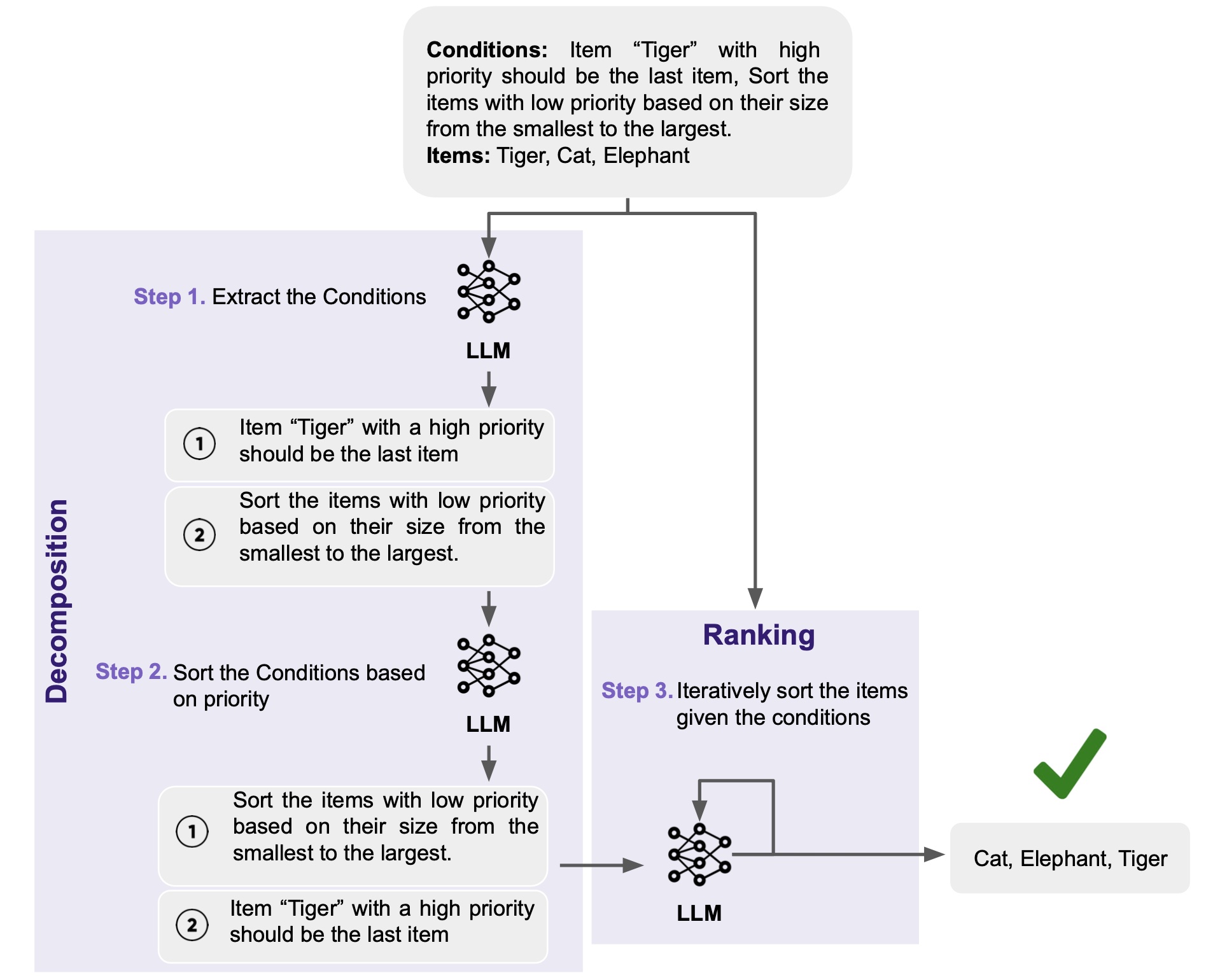}
        \caption{EXSIR (Ours)}
    \end{subfigure}  
    \caption{Overview of multi-conditional ranking. Instead of directly prompting LLMs to rank items based on the given conditions, we first extract and sort the conditions based on their priority. Then, we iteratively apply these sorted conditions to the item list.}
    \label{fig:mcr}
\minipostspace
\minipostspace
\end{figure*}
In this paper, we define and explore the task of multi-conditional ranking (MCR) by developing MCRank, a comprehensive benchmark consisting various item types and conditions for assessing MCR task. In addition, we also propose a novel decomposed reasoning based method, EXSIR, that beats strong baselines (including CoT) by up to 14.4\%. 
The new benchmark, 
MCRank,  spans a various category of conditions, including positional, locational, temporal, trait-based, and reasoning types. We have designed MCRank to address scenarios involving one to three conditions and to assess sets of 3, 5, or 7 items. The benchmark distinguishes between two types of items: token-level items, which consist of only a few tokens, and paragraph-level items, which contain up to 150 tokens. An example of MCRank involving two conditions and three token-level items is presented in Figure \ref{fig:mcr}.

Our initial investigations into the performance of existing LLMs on MCRank revealed a notable decline in accuracy as the number of items and conditions increased. Specifically, we observe that the accuracy of investigated LLMs , i.e., OpenAI o1-mini, GPT-4 \citep{openai2023gpt-4}, ChatGPT\citep{kocon2023chatgpt} (both turbo versions), Llama3.1-70B \citep{touvron2023llama} and Mistral (7B) \citep{jiang2023mistral} in correctly ranking items fell dramatically as the task scaled to three conditions and seven items, with accuracy approaching nearly 0\%.
To address the shortcomings of existing LLMs in MCR task, we introduce a novel method based on decomposed reasoning. Rather than directly prompting LLMs to rank items based on the given conditions, our approach begins with extracting and sorting the conditions based on their priority. Subsequently, we iteratively apply these sorted conditions to the item list. An illustration of our approach is provided in Figure \ref{fig:mcr}. Applying our method to MCRank, we observed a notable improvement, with up to a 14.4\% increase in the LLMs' ranking accuracy.

Observing the impact of our approach in improving LLMs performance on MCRank, we conduct an in-depth analysis of the models' performance, dissecting the results based on the types of items and conditions involved. Additionally, we examine the accuracy of the decomposition step within the evaluated LLMs shedding light on observed behavior of LLMs on MCRank. 
To delve deeper into the significance of the decomposition process, we incorporate a zero-shot chain-of-thought \citep{wei2022chain} approach, further underscoring the importance of segmenting the MCR task into multiple steps to achieve improved outcomes. 
Finally, we employ SFR \citep{meng2024sfrembedding}, ColBERT \citep{khattab2020colbert} and RankGPT (based on GPT-4) \citep{sun2023chatgpt}, rankers renowned for their performance in document ranking \citep{nguyen2016ms,dietz2017trec}, to represent existing rankers. Our comparison illustrates that, despite their success in existing ranking tasks, they exhibit considerably inferior performance on MCRank.

\section{Multi-Conditional Ranking}
The task of multi-conditional ranking is designed to shift the focus from traditional ranking tasks, which typically involve ordering a large set of items based on a single query. 
Instead, this task concentrates on sorting a smaller, pre-selected set of items according to multiple conditions. These conditions may not only conflict with one another but also carry varying levels of priority, adding layers of complexity to the ranking task. 
Moreover, each condition may specify a complete order for all items or only provide a partial ordering instructions for the placement of certain items. 
The primary goal is to simulate a scenario where a user provides a string containing several conditions along with their respective priorities, to guide the ranking of a already shortlisted group of items.
This task's complexity lies in balancing conditions, understanding their relative importance, and applying them to create a contextually relevant items order.

\begin{table}[t]
\small
\centering
    \begin{tabular}{c|l}
        \toprule
        \textbf{Type}&\textbf{Condition Examples}\\
        \midrule  
        Position& Item ``[X]'' should be the last from left\\
        \midrule 
        \multirow{2}{*}{Location}&Items that are in \textcolor{blue!80!black}{Africa} should appear at the\\
        &beginning\\
        \midrule
        \multirow{2}{*}{Temporal}& Sort items based on their \textcolor{blue!80!black}{deadline} from the\\
        &first to the last\\
        \midrule
        \multirow{2}{*}{Trait}&Sort the items based on their \textcolor{blue!80!black}{size} from the\\
        &smallest to the largest\\
        \midrule
        \multirow{2}{*}{Reason}&Items that has the largest \textcolor{blue!80!black}{yards of touchdown}\\
        &should appear at the beginning\\
        \bottomrule
    \end{tabular}
    \caption{Example of conditions based on different types. After extracting items, we determine their golden ranking based on their corresponding \textcolor{blue!80!black}{labels}. The conditions may specify a complete or partial order for item placement.}
    \label{tab:exp}
\minipostspace
\minipostspace
\end{table}
\subsection{MCRank Benchmark}
To develop a benchmark for assessing the ability of LLMs to tackle the multi-conditional ranking (MCR) task—where the goal is to rank a small set of items based on a string of unsorted conditions—we must first compile a collection of items, each tagged with a \emph{gold label} that denotes a specific category or a value for a particular feature. 
These labels serve as the foundation for generating the correct ranking order under any given set of conditions.
We structure the benchmark by classifying conditions into five types and categorizing items as either: (1) \emph{token-level}, which includes items comprising only a few tokens, and (2) \emph{paragraph-level}, which encompasses items containing up to 150 tokens. 
We then collect items and their corresponding labels for each category. 
The conditions are divided into the categories below. 
We aimed to be as comprehensive as possible in choosing the categories, drawing insights from previous works on various tasks such as question answering \citep{dua2019drop}, relation extraction \citep{pawar2017relation}, text classification \citep{pang2008opinion}, and retrieval systems \citep{zhao2024dense}. 
Our motivation for each category was to capture a broad range of ranking needs across various subfields. 
The detailed list of samples for each category and the datasets used is provided in the Appendix. We also provide an example for each condition category in Table \ref{tab:exp}. 

\vspace{1mm}
\noindent\textbf{Positional}: We define positional conditions as the ones that explicitly requests the placement of an item in a specific position within the ranking. Previous research \citep{srivastava2022beyond} has typically focused on straightforward conditions, such as positioning Item X in Position Y, where LLMs have demonstrated high levels of performance. However, in this work, we opt for more realistic and challenging conditions, such as ``Item X should be the last item from the left'' aiming to mimic situations where a user's objective is to modify the perceived importance of certain items by strategically placing them at either the end or the beginning of the list. This setting introduces a greater degree of complexity, requiring the model to interpret more nuanced spatial language and apply it accurately within the context of MCR task. This type of condition does not require pre-defined labels for the items.

\vspace{1mm}
\noindent\textbf{Locational}: Locational conditions are defined as conditions that require the placement of items in the ranking based on their geographical attributes. For the token-level category, we compile items and their respective locational labels by extracting popular entities and their objects from the \emph{location} predicate found in the T-REx benchmark  \citep{elsahar2018t}. 
Additionally, for the paragraph-level category, we combine prompts for \emph{birth place}, \emph{death place}, \emph{country of citizenship}, \emph{headquarter location}, and \emph{location} predicates from the T-REx benchmark, along with job descriptions from Dice\footnote{We extracted the data from \url{https://www.kaggle.com/datasets/PromptCloudHQ/us-technology-jobs-on-dicecom}} containing locational labels. 

\vspace{1mm}
\noindent\textbf{Temporal}: Temporal conditions are defined as conditions that dictate the placement of items based on their associated dates for specific attributes, such as birthdates. For the token-level category, we consider celebrities and their birthdates, sourced from the CACD benchmark \citep{chen2014cross}. For the paragraph-level category, we incorporate a mix of job description and their deadlines from Dice and paragraphs from SQUAD \citep{rajpurkar2016squad} that have a query about a publication date. 

\vspace{1mm}
\noindent\textbf{Trait-Based}: We characterize trait-based conditions as those that control the positioning of items predominantly based on a physical attribute. For the token-level category, we compile items along with their size and height information from the VEC benchmark \citep{li2023can}. Additionally, for the paragraph-level category, we consider Amazon reviews detailing attributes like size, color, and spice variety \citep{ni2019justifying, yang2022mave}, in addition to prompts derived from the \emph{genre} predicate in the T-REx benchmark. 

\vspace{1mm}
\noindent\textbf{Reason-Based}: We define reason-based conditions as those that necessitate logical/mathematical reasoning to determine the correct positioning of items, such as deducing the category of an item or performing mathematical operations on values of a certain attribute in each item. For the token-level category, we collect items and their categories from the auto-categorization task in Big-Bench \citep{srivastava2022beyond}. In the paragraph-level category, we sourced items from DROP's \citep{dua2019drop} paragraphs featuring "How Many" questions that require mathematical reasoning.

To develop the MCRank benchmark, we assemble a collection of datasets, each corresponding to one of the two item categories and featuring samples with 1, 2, or 3 conditions and sets of 3, 5, or 7 items, culminating in 18 distinct \emph{scenarios}. We curate the dataset for each scenario through several steps: Initially, for each condition type, we compile data and their labels to create 200 samples. Each sample includes a condition from that category, a randomly arranged set of items, and the correct item ranking based on the labels. For the positional type conditions, we utilize items from the auto-categorization task in Big-Bench for the token-level and Amazon reviews for the paragraph-level category.
After assembling 200 samples for each type of condition across all scenarios, we introduce additional conditions for scenarios requiring more than one condition to mimic a realistic setting where users specify various conditions. We randomly add either a condition to sort items based on character counts or a positional condition. In scenarios with three conditions, we incorporate both.

As in \cite{boutilier2013computational}, we consider the users' priority as extra input, and assume they are explicitly provided by the users. As described in \cite{schnabel2020impact,boutilier2013computational}, the aim is to discover enough about the user utility function to recommend a good recommendation. Thus, we further assign a ``\textbf{low priority}'' to the character count condition, a ``\textbf{medium priority}'' to each category type condition, and a ``\textbf{high priority}'' to the extra positional condition. 
The main usage of assigning priorities is to handle \textbf{contradictory} conditions. Moreover, our goal in choosing this specific assignment of conditions and their priorities is to capture the varying priorities a user might have in real-world contexts \cite{stray2024building}. Specifically, we use character count to represent low-priority and easy conditions that a user might inquire about. A medium-priority condition is covered by samples from different categories of conditions, while a high-priority condition represents a scenario in which a user asks to place the hardest/easiest, highest/lowest quality, or most/least qualified item either first (as most important) or last (to be ignored). We believe that this specific form of assignment, along with the 
randomness/diversity in selecting various conditions, can encompass a wide range of potential conditions posed by users in real-world scenarios.

To further validate our assignment of priorities, let's see a concrete application scenario for MCR in recommendation systems (more scenarios in the Appendix): Imagine a user interacting with a movie recommender system. After receiving a list of potential matches, the user may not want to read through the plot of each movie. Instead, based on their current mood, they may wish to rank the movies according to various conditions: (1) Movies should be sorted by their length, with low priority, as the user generally prefers shorter films but doesn't mind longer ones either; (2) The movies should be sorted by their IMDB score, with medium priority; and (3) The highest-rated movie should be placed at the end of the list, with high priority, because the user has already watched it.

Upon assigning the priorities, we randomize the conditions order, adding another layer of complexity to make the task more realistic, and combine the samples from each condition type to form a dataset for each scenario. To maintain clarity, we eliminate samples where multiple items share the same character counts. We curated approximately 1000 samples for each scenario. We provide MCRank's statistics, more details, and a step by step illustration of MCRank creation in the Appendix.

\subsection{Extracting, Sorting, and Iteratively Ranking (EXSIR)}
As shown in Section \ref{sec:rank}, the performance of current LLMs when tested on the MCRank benchmark reveals a pronounced decline, particularly noticeable as the complexity of the task increases with additional conditions and items. To address this challenge and improve the LLMs' effectiveness, we introduce a novel strategy based on the decomposed reasoning method, which meticulously breaks down the multi-conditional ranking task into several manageable steps. 

The process begins with the extraction of individual conditions from a given string, organizing these into a coherent list. Following this, we implement a sorting mechanism that arranges the conditions based on their assigned priorities. This prioritization is crucial for the subsequent step, where these sorted conditions are iteratively applied to the list of items. In this phase, the item list is iteratively updated, with each cycle refining the rankings based on the current condition being applied. To illustrate this process, we provide a visual representation in Figure \ref{fig:mcr}, which outlines the EXSIR method's workflow. Consistency in our approach is maintained by using the same LLM across all steps of the EXSIR process. To offer further clarity and insight into our methodology, detailed descriptions of the prompts utilized at each stage are available in the Appendix. 

\begin{figure*}[t!]
    \centering
    \begin{subfigure}[b]{0.5\textwidth}
        \centering
        \includegraphics[width=\linewidth]{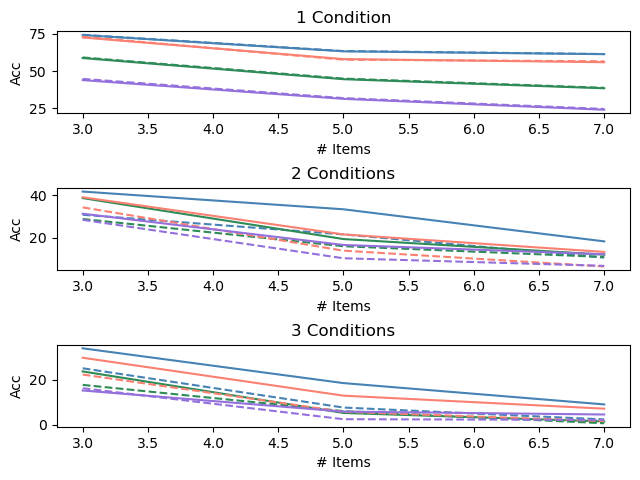}
        \vskip -1mm
        \caption{Accuracy}
    \end{subfigure}%
    ~ 
    \begin{subfigure}[b]{0.5\textwidth}
        \centering
        \includegraphics[width=\linewidth]{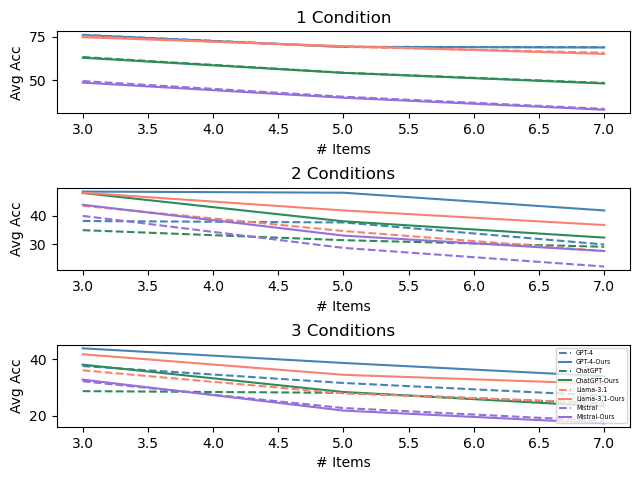}
        \vskip -1mm
        \caption{Average Accuracy}
    \end{subfigure}
    \vskip -2mm
    \caption{LLMs performance on MCRank for token-level items.}
    \label{fig:rank-token}
\minipostspace
\minipostspace
\end{figure*}
\section{Experimental Details}
\noindent\textbf{Models}: We evaluate both commercial LLMs, OpenAI o1-mini, GPT-4 \citep{openai2023gpt-4}, ChatGPT\citep{kocon2023chatgpt} (both turbo versions), as well as open-source LLMs, Llama3.1-70B (instruct) \citep{touvron2023llama} and Mistral (Mistral-7B-Instruct-v0.2) \citep{jiang2023mistral} on MCRank. To address the MCR task, in the base setting, we input the string of conditions along with the list of items into the prompt, instructing the LLMs to organize the list according to the conditions. For paragraph-level items, we assign a unique label, "Item-K," to each item. The task for the model is then to rank the items but to output the sequence of sorted labels---Item-K---instead of the items themselves. The details of all prompts utilized in this study are provided in the Appendix. 
We designed the prompts to mirror previous ranking works, while avoiding framing the task as a standard document monotonic-ranking task.

\vspace{1mm}
\noindent\textbf{Evaluation Metric}:
Given that the MCR task defined in this paper represents a broader and more complex variation of previously defined ranking tasks---where, unlike those tasks, the significance or relevance of items in the gold ranking doesn't necessarily diminish in a linear order---traditional ranking metrics like MRR or nDCG \citep{zangerle2022evaluating} are not suitable for our context. Consequently, we evaluate model performance on the MCR task using \textbf{exact match accuracy},  scoring 1 for a fully correct ranking and 0 for an incorrect one. Additionally, we employ an \textbf{averaged accuracy} metric, measuring the mean number of correctly ranked items per sample for a more nuanced assessment of models.

\section{Experiments}
We explore the effectiveness of LLMs and the impact of EXSIR on the MCR task using the MCRank benchmark, starting with an evaluation of models' performances. 
Subsequently, we provide per-category breakdown of performances to assess the impact if each condition category. We assess the accuracy of the decomposition step for each model to gauge its impact on EXSIR's functionality. Lastly, to underscore the importance of decomposed reasoning through multiple steps, we compare our method's performance against zero-shot CoT prompting and existing ranking models.

\subsection{Ranking on MCRank Benchmark}
\label{sec:rank}
\begin{figure*}[t!]
    \centering
    \begin{subfigure}[b]{0.5\textwidth}
        \centering
        \includegraphics[width=\linewidth]{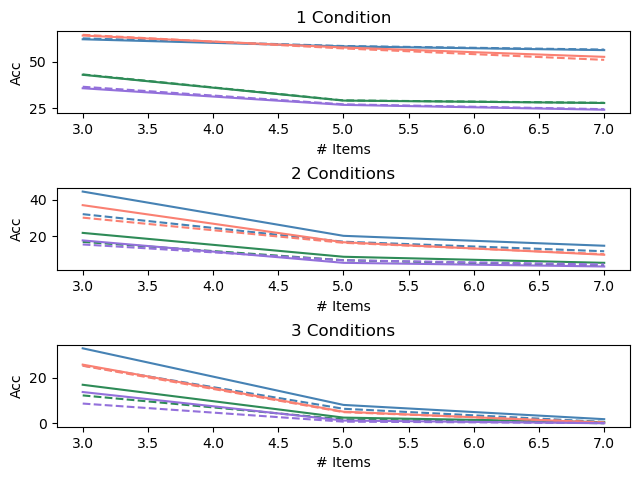}  
        \vskip -1mm
        \caption{Accuracy}
    \end{subfigure}%
    ~ 
    \begin{subfigure}[b]{0.5\textwidth}
        \centering
        \includegraphics[width=\linewidth]{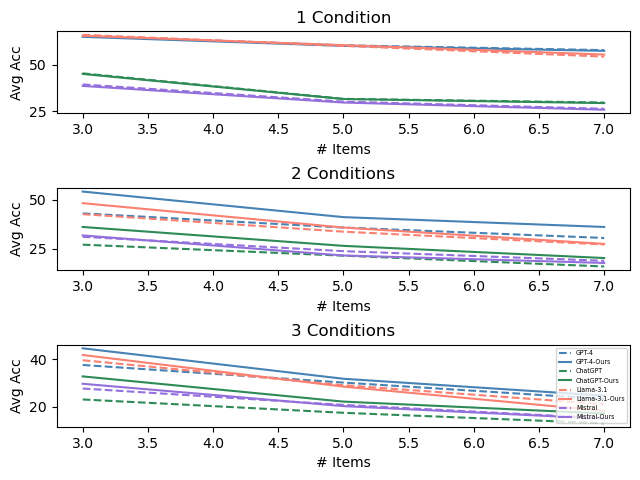}
        \vskip -1mm
        \caption{Average Accuracy}
    \end{subfigure}
    \vskip -2mm
    \caption{LLMs performance on MCRank for paragraph-level items.}
    \label{fig:rank-para}
\minipostspace
\minipostspace
\end{figure*}

The accuracy and average accuracy of LLMs with and without EXSIR are depicted in Figures \ref{fig:rank-token} and \ref{fig:rank-para}. 
These figures reveal that while all evaluated LLMs exhibit significant accuracy with a single condition and three items, their performance rapidly declines towards zero as the number of conditions rises to three and the items to seven. In overall, there is a consistent pattern observed between token-level and paragraph-level items, where a noticeable decrease in performance occurs as we transition from the token to the paragraph setting.

Notably, EXSIR significantly and consistently enhances model performance across various settings, with the most pronounced improvement observed in GPT-4, likely due to its superior performance in the decomposition step (further discussed in Section \ref{sec:dec}). Additionally, a similar trend is noticeable in both accuracy and average accuracy across the models. Intriguingly, despite the convergence of accuracy to zero in more complex scenarios, the average accuracy remains substantial, highlighting the fragility of accuracy in the MCR setting which is aligned with previous works on the impact of metrics sensitivity in LLMs performance on reasoning tasks \citep{schaeffer2024emergent}. 

\subsection{Per-Category Breakdown}
A detailed breakdown of LLMs' performance across different condition categories on the MCRank benchmark is available in the Appendix. 
Notably, performance varies across condition categories for each setting. In token-level scenarios, models excel in reason-based conditions, whereas, in paragraph-level settings, they perform better in locational conditions but exhibit comparable results in trait-based conditions. 
This variation is due to the increased complexity of reason-based conditions in paragraph settings and the explicit provision of label information in trait-based and locational conditions, simplifying these tasks. 
However, all models struggle with the positional conditions, especially with conditions like ``item [X] should be the last from the right'' and ``the last item in the sorted list should appear in the first place.'' This struggle is likely due to the conflict between the conditions and the models' autoregressive nature, which necessitates a full understanding of the final ranking before even generating the first item. We also provide the accuracy of LLMs in satisfying high priority conditions in Appendix.

\subsection{Accuracy of Decomposition}
\label{sec:dec}
Now that we have seen how EXSIR improve LLMs' performance, one remaining question is that how the accuracy of the decomposition step influences the overall performance. 
We detail the accuracy of LLMs in extracting and sorting conditions (decompostion step) in Table \ref{tab:con-acc}. The result indicates that GPT-4 outperforms other LLMs, whereas Mistral's accuracy decreases when transitioning to paragraph-level, correlating with its EXSIR-augmented ranking performance in such scenarios. 

\begin{table}[t]
\small
\centering
\begin{tabular}{lrrrr}
\toprule 
\multirow{2}{*}{\bf Models} & \multicolumn{2}{c}{\bf Token}&  \multicolumn{2}{c}{\bf Paragraph}\\
\cmidrule(lr){2-3}
\cmidrule(lr){4-5}
&2 cond&3 cond&2 cond&3 cond\\
\midrule
Mistral&82.9&81.5&70.3&66.6\\
Llama3.1&91.0&87.2&88.1&86.0\\
ChatGPT&83.1&79.6&82.3&79.5\\
GPT-4&97.3&96.7&91.3&85.6\\
\bottomrule
\end{tabular}
\caption{LLMs accuracy in extracting and sorting the conditions (decomposition part).}
\label{tab:con-acc}
\postspace
\minipostspace
\end{table}
\begin{figure*}[t!]
    \centering
    \includegraphics[width=\linewidth]{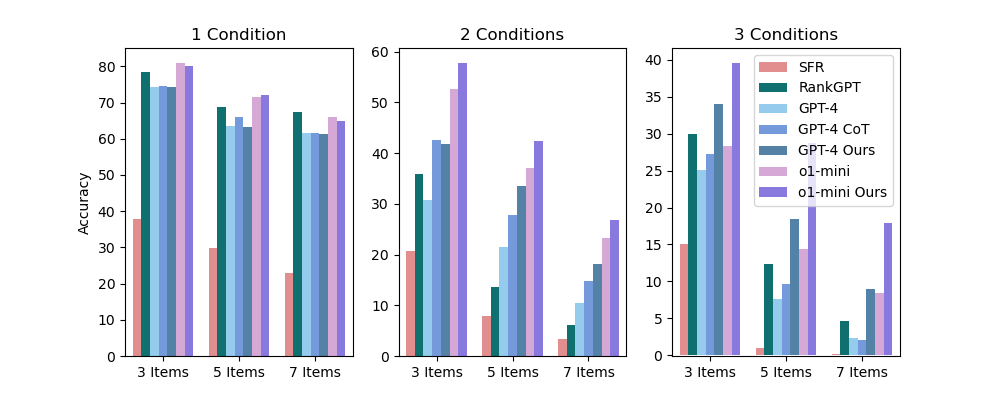}
    \vskip -1.8mm
    \caption{Evaluating the impact of EXSIR against zero-shot CoT prompting for token-level items. We additionally report SFR and RankGPT performances as representatives of existing rankers.}
    \label{fig:tok-bar}
\minipostspace
\minipostspace
\end{figure*}

These findings, coupled with the LLMs' performances in previous sections, suggest that for EXSIR to significantly influence model performance, a high accuracy in the decomposition step is crucial, along with at least an adequate performance in ranking items under a single condition. 
Consequently, to enhance ranking performance while maintaining the efficiency of open-source models, one potential strategy could be employing a more advanced model like GPT-4 for the decomposition step and utilizing less powerful models for the ranking process. The investigation of such strategies is a promising avenue for future research.

\subsection{Zero-shot CoT vs Decomposed Reasoning}
So far, we have seen how EXSIR enhances LLMs' performances. However, one might ask: is multi-step decomposition necessary, or could similar results be achieved by combining the steps into a single prompt, like the zero-shot Chain-of-Thought (CoT) approach \citep{wei2022chain}? This section narrows the focus to GPT-4, comparing its performance using EXSIR against zero-shot CoT-style prompting. We provide an example of a CoT-based prompt and discuss the impact of CoT prompting on Llama3.1's performance, further reinforcing our findings with GPT-4 in the Appendix.

\begin{figure*}[t!]
    \centering
    \includegraphics[width=\linewidth]{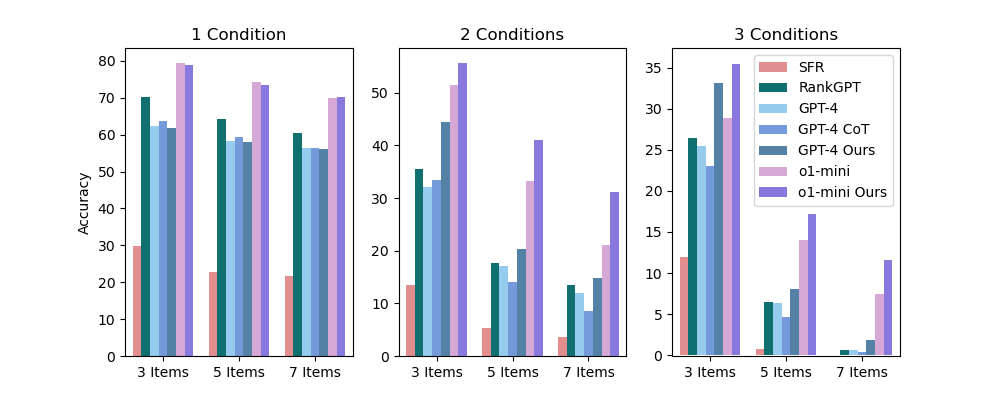}
    \vskip -1.8mm
    \caption{Evaluating the impact of EXSIR against zero-shot CoT prompting for paragraph-level items. We additionally report SFR and RankGPT performances as representatives of existing rankers.}
    \label{fig:para-bar}
\minipostspace
\minipostspace
\end{figure*}

The accuracy of GPT-4, GPT-4 with CoT and with EXSIR on MCRank, is depicted in Figures \ref{fig:tok-bar} and \ref{fig:para-bar}. For token-level items, the figures demonstrate that while CoT prompting boosts the base performance of GPT-4, there remains a notable performance disparity between EXICR and CoT, highlighting the value of multi-step reasoning. In contrast, for paragraph-level items, incorporating CoT instructions seems to decrease the base model's performance, possibly due to the task complexity and the challenge of adhering to the provided CoT instructions for GPT-4 (we provide more discussion on the impact of few-shot CoT in Appendix).

\subsection{Existing Rankers and o1-mini}
In here, we first evaluate the performance of existing rankers, followed by an assessment of the reasoning-focused LLM o1-mini on MCRank. 

\paragraph{Existing rankers:} We evaluate the performance of SFR \citep{meng2024sfrembedding} and ColBERT \citep{khattab2020colbert}, as representative encoder-based ranking models, along with RankGPT (based on GPT-4) \citep{sun2023chatgpt} on MCRank, due to their strong performance on existing benchmarks. 
The accuracy of SFR and RankGPT on MCRank are illustrated in Figures \ref{fig:tok-bar} and \ref{fig:para-bar} (the performance of ColBERT, along with implementation details, is provided in the Appendix). These figures indicate that SFR's performance is significantly inferior compared to GPT-4 (and most of other investigated LLMs), highlighting the complexity of the task and the limitations of smaller ranking models in a multi-conditional setting. Moreover, since RankGPT is designed to monotonically rank documents based on a given query, it outperforms the GPT-4 baseline when only one condition is present. However, as the number of conditions increases, its performance becomes comparable to our vanilla GPT-4 baseline. Based on these results, integrating RankGPT into EXSIR's ranking step could improve performance, which we leave for future research.

\vspace{1mm}
\noindent\textbf{o1-mini:} To further evaluate the complexity of MCRank, we present the performance of the reasoning-focused LLM o1-mini in Figures \ref{fig:tok-bar} and \ref{fig:para-bar} (further experimental results on o1-mini's are provided in the Appendix). 
Despite its strong performance on other reasoning tasks, o1-mini shows relatively poor performance on the MCRank benchmark. As evident, GPT-4 combined with EXSIR outperforms o1-mini in many cases. However, in the paragraph-level setting, in cases where more items are involved, o1-mini shows an advantage and performs better. Moreover, as we can observe, applying EXSIR on top of o1-mini significantly improves performance, further demonstrating the effectiveness of the EXSIR approach.
\section{Related Work}
LLMs have achieved significant success in ranking tasks,  
but most efforts focus on ranking passages based on a single query, despite the broad real-world applications of ranking.

\vspace{1mm}
\noindent\textbf{Ranking with LLMs}:
In recent years, LLMs have become pivotal in addressing ranking related tasks. 
Initially focusing on encoder-based rankers \citep{nogueira2019multi,khattab2020colbert}, the rapid advancement of autoregressive LLMs has led to the development of methodologies that utilize these models as rankers, achieving unparalleled performance across various benchmarks \citep{zhuang2023open, qin2023large}. However, despite these advances, the majority of LLM-based ranking efforts have concentrated on ordering extensive lists of passages based on a query, often overlooking the diverse applications of ranking in real-world scenarios. 
Our work closely aligns with developments in recommendation systems, such as the conditional methods proposed by \citet{hou2024large}, which only considers a limited concept of condition in regard to variety and complexity compared to our notion of multi-conditional ranking.

\vspace{1mm}
\noindent\textbf{Decomposed Reasoning with LLMs}:
As LLMs grow stronger, decomposed reasoning has emerged as a fundamental strategy to enhance their capabilities by segmenting complex tasks into smaller, more manageable components. 
This decomposition can be straightforward, utilizing a single LLM, as seen in approaches like Chain-of-Thought \citep{wei2022chain}, Tree-of-Thought \citep{yao2024tree}, and Self-Verification \citep{weng2022large}. Alternatively, it can involve more complex interactions among multiple models within a multi-agent systems \citep{xi2023rise, guo2024large}. Prior research has successfully integrated decomposed reasoning into various tasks, including question answering \citep{dua2022successive}, retrieval-augmented generation (RAG) \citep{asai2023self}, and mathematical reasoning \citep{qi2023art}.

\section{Conclusion}
We explore the multi-conditional ranking task, an essential but underexplored aspect of ranking in real-world applications. Introducing MCRank benchmark, we have highlighted the challenges LLMs face when ranking a small set of items under a variety of complex and sometimes conflicting conditions. Our investigation reveals a significant performance drop in LLMs as the number of conditions and items increases. To address this, we introduce EXSIR, a novel decomposed reasoning method that improves LLM performance by up to 14.4\% in accuracy. 
We also analyze the performance of LLMs across various condition categories and the effectiveness of the decomposition step in enhancing accuracy. 
Finally, by contrasting our approach with other existing methodologies such as CoT and existing rankers, we demonstrate its effectiveness and the complexity of the MCR task.

\section{Limitations}

While this study advances our understanding of multi-conditional ranking with LLMs, several limitations should be take into consideration:

\vspace{1mm}
\noindent\textbf{Limited Scope of LLMs:} Our research focused on five specific LLMs and three existing rankers, which, while prominent, do not encompass the full spectrum of models available in the field. This narrow focus may not fully capture the diversity of capabilities present in the broader landscape of large language models.

\vspace{1mm}
\noindent\textbf{Model Type Restriction:} We limited our exploration to autoregressive models and encoder-type rankers. Potentially, encoder-decoder models, known for their robust performance in a variety of NLP tasks, might exhibit different behaviors and capabilities when applied to the MCR task. We leave the exploration of these type of LLMs to future research.

\vspace{1mm}
\noindent\textbf{EXSIR Efficiency:} Our proposed method EXSIR, while effective in enhancing performance on MCR tasks, presents notable challenges in terms of efficiency and cost. As a multi-step ranking process, EXSIR inherently is more time-consuming and costly compared to simpler, single-step methods. This issue becomes more pronounced when deploying EXSIR at scale in real-world applications. Optimizing the efficiency of EXSIR, without compromising its performance benefits, remains an open area for future research.

\vspace{1mm}
\noindent\textbf{Single LLM for Decomposition and Ranking:} In our methodology, the same LLM is used for both the decomposition and ranking steps. This approach might not be optimal, as different models could have varying strengths, with some excelling at decomposition and others at ranking. A more nuanced strategy could involve a multi-agent system, where a planner identifies and decomposes the conditions, and then divide the ranking tasks to different rankers based on each condition. This division of labor could enhance the overall effectiveness of the multi-conditional ranking process.

\vspace{1mm}
\noindent\textbf{Interactive Ranking Solution:} Our current model does not incorporate user interaction, which could be a significant limitation. An interactive ranking system, where the user engages in a dialogue with the system to refine the ranking iteratively, could offer a more dynamic and user-tailored solution. This approach would allow the system to adapt to user feedback in real-time, potentially leading to more accurate and satisfactory ranking outcomes.

Addressing these limitations in future work could broaden our understanding of multi-conditional ranking, improve the performance and applicability of ranking systems, and offer a more nuanced perspective on the integration of LLMs in such tasks.

\bibliography{custom}

\appendix

\section{MCRank Details}
In this section, we first outline the specifics of the datasets from which we extract items' label to construct MCRank. Subsequently, we present more details on how we create MCRank and provide a comprehensive list of the various conditions included in the MCRank benchmark.

\subsection{Benchmark Used To Create MCRank}
For token-level items, we utilized the following datasets: The T-REx benchmark \citep{elsahar2018t}, which includes a subset of Wikipedia triples aligned with corresponding Wikipedia abstracts, featuring a comprehensive collection of 11 million triples and 3.09 million Wikipedia abstracts across more than 600 unique Wikidata predicates. The CACD benchmark \citep{chen2014cross}, which comprises images and details such as the birthdate of 2,000 celebrities. The VEC benchmark \citep{li2023can}, designed to test LLMs' understanding of visual and embodied concepts, provides physical attributes like size and height for a range of entities. Additionally, the auto-categorization task in Big-Bench \citep{srivastava2022beyond} involves predicting the category to which a given list of items belongs. 

For paragraph-level items, along with T-REx, we incorporated the following datasets: A collection of 4.6 million job descriptions from Dice\footnote{The data was extracted from \url{https://www.kaggle.com/datasets/PromptCloudHQ/us-technology-jobs-on-dicecom}}, each detailing various attributes such as work location and application deadline. The SQUAD dataset \citep{rajpurkar2016squad}, a reading comprehension dataset composed of questions based on Wikipedia articles, where each question's answer is a text segment from the related passage. We also utilized Amazon reviews that contain attributes such as size, color, and spice variety \citep{ni2019justifying, yang2022mave}. Additionally, we used the DROP dataset \citep{dua2019drop}, a more complex reading comprehension dataset, where many questions necessitate reasoning about the information in the corresponding passage to find the answer.

\subsection{MCRank Creation Details}
\begin{figure*}[t!]
    \centering
    \includegraphics[width=\linewidth]{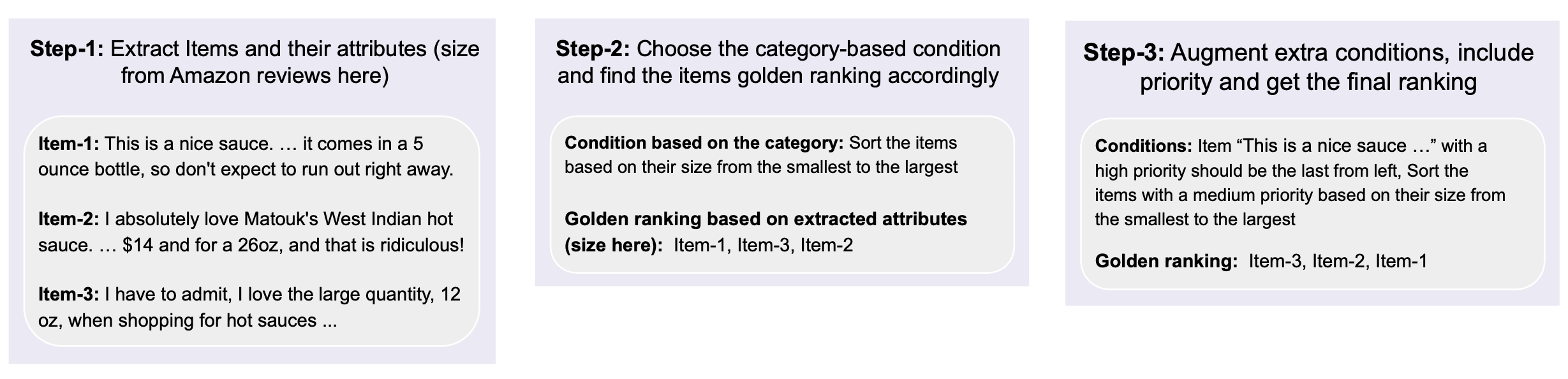}
    \caption{Step-by-step illustration of MCRank creation.}
    \label{fig:illust}
\minipostspace
\minipostspace
\end{figure*}

We provided a step-by-step illustration of MCRank creation in Figure \ref{fig:illust}. For each scenario in MCRank, we began with 200 samples for each category, utilizing extracted attributes from the original datasets to establish the golden ranking of items. Additional conditions were then applied on top of each category-based condition, necessitating a recalculation of the gold ranking to accommodate these augmented conditions. Throughout this process, we removed some samples where the addition of extra conditions resulted in non-unique golden rankings. Consequently, the average number of samples per category in MCRank ranges from 159 to 200. The statistics of MCRank are presented in Table \ref{tab:stat}.

\begin{table}
\small
\centering
\begin{tabular}{lrrr}
\toprule 
 & 1 Condition & 2 Conditions & 3 Conditions\\
\midrule
T-level &916.7&860.0&797.7\\
P-level&1000&1000&1000\\
\bottomrule
\end{tabular}
\caption{Average number of samples in MCRank benchmark per number of conditions for paragraph- (P-level) and token-level (T-level) items.}
\label{tab:stat}
\postspace
\minipostspace
\end{table}
\subsection{Conditions in MCRank}
The detailed list of conditions included in the MCRank benchmark is presented in Table \ref{tab:conds}.

\begin{table*}[t]
\small
\centering
    \begin{tabular}{c|p{11cm}}
        \toprule
        \textbf{Type}&\textbf{Conditions}\\
        \midrule  
        \multirow{6}{*}{\rotatebox[origin=c]{90}{\bf Position}}&1) Item ``[X]'' should be the last from left\\
        &2) Item ``[X]'' should be the last from right\\
        &3) First item in the final sorted order should appear in the end\\
        &4) First item in the final sorted order should appear in the beginning\\
        &5) Last item in the final sorted order should appear in the end\\
        &6) Last item in the final sorted order should appear in the beginning\\
        \midrule
        \multirow{4}{*}{\rotatebox[origin=c]{90}{\bf Location}}&7) Items that are in ``[X]'' should appear at the beginning\\
        &8) Items that are in ``[X]'' should appear at the end\\
        &9) Items that have ``[Y]'' in ``[X]'' should appear at the beginning\\
        &10) Items that that have ``[Y]'' in ``[X]'' should appear at the end\\
        \midrule
        \multirow{9}{*}{\rotatebox[origin=c]{90}{\bf Temporal}}&11) Sort the items based on their birthday from the oldest to the newest\\
        &12) Item that born before ``[X]'' should appear at the end\\
        &13) Item that born after ``[X]'' should appear at the beginning\\
        &14) Sort items based on their deadline from the first to the last\\
        &15) Item that has a deadline before ``[X]'' should appear at the end\\
        &16) Item that has a deadline after ``[X]'' should appear at the beginning\\
        &17) Sort items based on mentioned publication date from the first to the last\\
        &18) Item that has a publication date before ``[X]'' should appear at the end\\
        &19) Item that has a publication date after ``[X]'' should appear at the beginning\\
        \midrule
        \multirow{12}{*}{\rotatebox[origin=c]{90}{\bf Trait}}&20) Sort the items based on their size from the smallest to the largest\\
        &21) Sort the items based on their height from the shortest to the tallest\\
        &22) Item with a size of less than ``[X]'' should appear at the end\\
        &23) Item with a size of more than ``[X]'' should appear at the beginning\\
        &24) Sort the items based on their size from the smallest to the largest\\
        &25) Item that is a ``[X]'' should appear at the end\\
        &26) Item that is a ``[X]'' should appear at the beginning\\
        &27) Item with a ``[X]'' color should appear at the end\\
        &28) Item with a ``[X]'' color should appear at the beginning\\
        &29) Item with the ``[X]'' genre should appear at the end\\
        &30) Item with the ``[X]'' genre should appear at the beginning\\
        &31) Sort the items based on their character count from the smallest to largest\\
        \midrule
        \multirow{5}{*}{\rotatebox[origin=c]{90}{\bf Reason}}&32) Items in the category ``[X]'' should appear at the beginning\\
        &33) Items in the category ``[X]'' should appear at the end\\
        &34) Sort items based on ``[X]'' from the smallest to the largest\\
        &35) Items that has the largest ``[X]'' should appear at the beginning\\
        &36) Items that has the smallest ``[X]'' should appear at the end\\
        \bottomrule
    \end{tabular}
    \caption{List of conditions in MCRank. For instance, in location-based conditions, ``[Y]'' could represent ``country of citizenship''. In trait-based conditions, ``[X]'' might denote ``Spice Variety''. Similarly, in reason-based conditions, ``[X]'' could exemplify ``longest yards of touchdown''.}
    \label{tab:conds}
\end{table*}
\section{Real-World Example Scenarios for MCR}
In addition to the provided scenario for the use of MCR in recommendation systems, multi-conditional ranking can have many use-cases in other applications such as:

\paragraph{Education:} Consider a teacher who needs to select questions from a pre-existing list. Without needing to read or solve each question, the teacher might want to rank them based on conditions such as: (1) Questions from a specific topic (Topic X) should appear higher in the list, with low priority; (2) The questions should be sorted by difficulty, with medium priority; and (3) The most difficult questions should be placed at the end, with high priority, as the teacher prefers to reserve them for advanced students.

\paragraph{HR:} A recruiter is tasked with selecting a candidate from a list of already short-listed resumes. Without delving deeply into every resume, and spending a significant amount of time, the recruiter wants to rank the candidates according to conditions such as: (1) Candidates who graduated before 2015 should be lower on the list, with low priority; (2) Candidates with more years of NLP experience should be ranked higher, with medium priority; and (3) The most over-qualified candidate should appear at the end, with high priority, as the recruiter wants to focus on other candidates first.

\section{Details of Prompts}
The prompts utilized for ranking token-level and paragraph-level items are detailed in Prompts \ref{prompt:token} and \ref{prompt:paragraph}, respectively. Additionally, the prompts employed for the extraction and sorting of conditions are outlined in Prompts \ref{prompt:extract} and \ref{prompt:sort}. Finally, we provide an example of zero-shot CoT-based prompt for token-level items in Prompt \ref{prompt:token-cot}.

\begin{prompt}[title={\footnotesize\texttt{Token-level Ranking Prompt}}, label=prompt:token]
Given following conditions: ``[string of conditions]'', sort the list of items ``[string of items]'' from left to right. Do not provide any explanation.
\end{prompt}

\begin{prompt}[title={\footnotesize\texttt{Paragraph-level Ranking Prompt}}, label=prompt:paragraph]
Given following conditions: ``[string of conditions]'', sort the items from left to right. Do not provide any explanation and only provide a permutation of Item-1, ..., Item-k enter separated as the output.\\
Item-1: [item-1]\\
Item-2: [item-1]\\
...
\end{prompt}

\begin{prompt}[title={\footnotesize\texttt{Condition Extracting Prompt}}, label=prompt:extract]
Given the conditions, extract the conditions into numbered items separated by enter. Do not provide any explanation and do not modify the conditions.\\ 
Conditions: [string of conditions]
\end{prompt}

\begin{prompt}[title={\footnotesize\texttt{Condition Sorting Prompt}}, label=prompt:sort]
Given the conditions, sort these conditions in the order that they should be applied to a list of items sequentially based on their priority to satisfy all their requirements as much as possible from the lowest priority to the highest priority. Do not provide any explanation and do not modify the conditions. \\
Conditions: [list of extracted conditions]
\end{prompt}

\section{Detailed Breakdown of Ranking performance on MCRank}
A detailed breakdowns of models performance on MCRank, segmented by the category of conditions and items, are presented in Tables \ref{tab:token-1}, \ref{tab:token-2}, \ref{tab:token-3}, \ref{tab:para-1}, \ref{tab:para-2}, and \ref{tab:para-3}.

\section{LLMs Performances in Satisfying High Priority Conditions}
To better understand the relationship between LLMs' ranking performance and different provided conditions, we calculated the percentage of samples where the ``high priority'' condition was satisfied. Focusing on high-performing models in MCRank, i.e., OpenAI o1-mini, GPT-4, and Llama3.1, we present the results in Table \ref{tab:con-high-acc}. As shown, the accuracy of satisfying the ``high priority'' condition changes only slightly as the number of items increases. Additionally, LLMs enhanced with EXSIR show consistently higher accuracy, consistent with their exact match and average ranking accuracy. Notably, the accuracy of satisfying the 'high priority' condition is significantly higher—especially in GPT-4 and Llama3.1—compared to exact match ranking accuracy, suggesting this condition is only partially responsible for the overall low performance of LLMs on MCRank.

\begin{table*}[t]
\small
\centering
\begin{tabular}{lrrrrrr}
\toprule 
\multirow{2}{*}{\bf Models} & \multicolumn{3}{c}{\bf Token-level}&  \multicolumn{3}{c}{\bf Paragraph-level}\\
\cmidrule(lr){2-4}
\cmidrule(lr){5-7}
&3 items&5 items&7 items&3 items&5 items&7 items\\
\midrule
o1-mini&34.0&31.1&31.0&38.6&33.1&33.0\\
o1-mini-Ours&51.6&51.2&47.9&55.2&52.8&49.4\\
GPT-4&39.4&36.2&35.7&41.3&39.3&36.0\\
GPT-4-Ours&50.7&50.5&48.3&54.4&47.4&44.9\\
Llama3.1&37.0&33.6&33.4&43.7&37.4&35.7\\
Llama3.1-Ours&51.1&49.6&49.1&45.5&43.7&37.1\\
\bottomrule
\end{tabular}
\caption{LLMs' accuracy in satisfying high priority conditions when ranking items in MCRank.}
\label{tab:con-high-acc}
\minipostspace
\minipostspace
\end{table*}

\section{Impact of Zero-Shot CoT on Llama3.1}
The accuracy of Llama3.1, Llama3.1 with zero-shot CoT, and Llama3.1 with EXSIR on MCRank is shown in Figures \ref{fig:llama-tok-bar} and \ref{fig:llama-par-bar}. As shown, the figures reveal that while CoT prompting, in most cases, improves Llama3.1's base performance, there remains a significant performance gap, especially as the number of conditions increases, between EXSIR and CoT, emphasizing the benefits of multi-step reasoning. 

\section{Impact of Few-Shot CoT on MCRank}
While we only present zero-shot CoT results in the paper, we did investigate the performance of few-shot CoT prompting on the MCR task. Our observations showed that few-shot CoT performed only marginally better, and only when the provided demonstrations were fully aligned with the target sample—specifically, when they had the same number and category of conditions, as well as a similar number and type of items (consistent with similar observations reported in \citep{stechly2024chain}). Moreover, since assuming access to demonstrations that perfectly mirror the target sample is unrealistic in real-world scenarios, and to ensure a more consistent evaluation across various cases, we chose to focus primarily on zero-shot CoT results in the paper.

\section{More Details on Existing Rankers and o1-mini}
\begin{figure*}[t!]
    \centering
    \includegraphics[width=\linewidth]{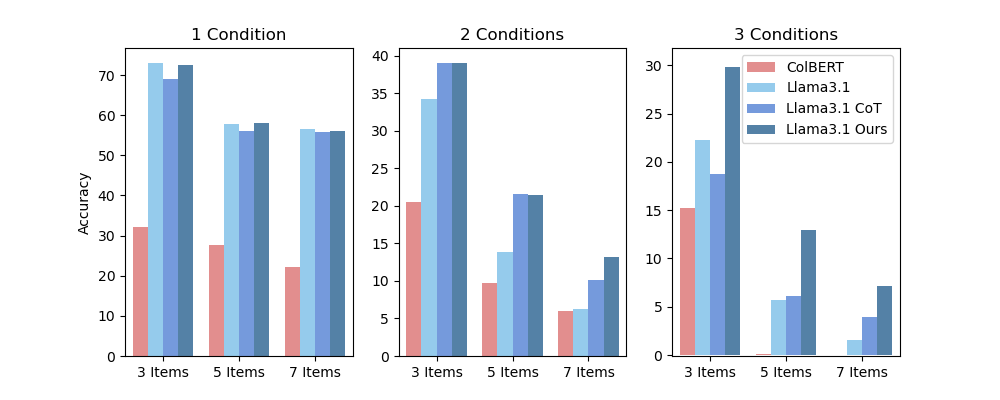}
    \caption{Evaluating the impact of EXSIR on Llama3.1 against zero-shot CoT prompting for token-level items. We additionally report ColBERT performance.}
    \label{fig:llama-tok-bar}
\minipostspace
\minipostspace
\end{figure*}

\begin{figure*}[t!]
    \centering
    \includegraphics[width=\linewidth]{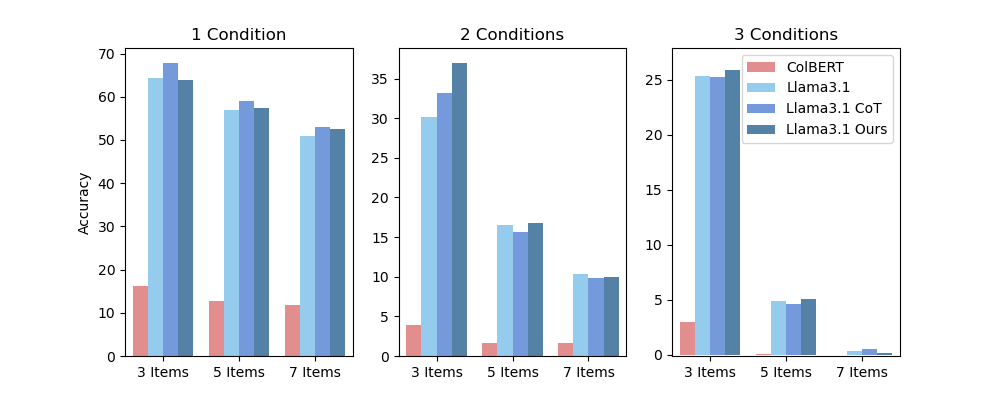}
    \caption{Evaluating the impact of EXSIR on Llama3.1 against zero-shot CoT prompting for paragraph-level items. We additionally report ColBERT performance.}
    \label{fig:llama-par-bar}
\minipostspace
\minipostspace
\end{figure*}

To evaluate the performance of encoder-based rankers on MCRank, we treat the instruction string as the query. Then, using LlamaIndex \citep{Liu_LlamaIndex_2022}, we rank the list of items for each sample individually. The results for ColBERT on MCRank are shown in Figures \ref{fig:llama-tok-bar} and \ref{fig:llama-par-bar}. We also provided more detailed results of and o1-mini performance in Tables \ref{tab:o1-mini-tok} and \ref{tab:o1-mini-par}. Moreover, the average accuracy of decomposition step for o1-mini (similar to the results in Table 3) are as follows: for token-level items, 98.4\% with 2 conditions and 95.0\% with 3 conditions; for paragraph-level items, 96.1\% with 2 conditions and 93.2\% with 3 conditions.

\begin{table*}
\small
\centering
\begin{tabular}{llrrrrrr}
\toprule 
&\multirow{2}{*}{\bf Models}&\multicolumn{2}{c}{\bf 3 items}&\multicolumn{2}{c}{\bf 5 items}&\multicolumn{2}{c}{\bf 7 items}\\
\cmidrule(lr){3-4}
\cmidrule(lr){5-6}
\cmidrule(lr){7-8}
&&ACC&Avg ACC&ACC&Avg ACC&ACC&Avg ACC\\
\midrule
\multirow{3}{*}{\rotatebox[origin=c]{90}{\bf Base}}&1 condition&81.0&81.7&71.4&74.9&66.1&74.5\\
&2 conditions&52.6&56.4&37.1&50.0&23.2&40.1\\
&3 conditions&28.3&38.2&14.4&34.0&8.4&31.8\\
\midrule
\multirow{3}{*}{\rotatebox[origin=c]{90}{\bf Ours}}&1 condition&80.2&81.3&72.1&75.6&65.0&73.8\\
&2 conditions&57.8&61.1&42.3&54.9&26.8&46.2\\
&3 conditions&39.6&47.1&28.8&43.2&17.9&37.6\\
\bottomrule
\end{tabular}
\caption{o1-mini and o1-mini with EXSIR (ours) performances for token-level setting in MCRank.}
\label{tab:o1-mini-tok}
\end{table*}

\begin{table*}
\small
\centering
\begin{tabular}{llrrrrrr}
\toprule 
&\multirow{2}{*}{\bf Models}&\multicolumn{2}{c}{\bf 3 items}&\multicolumn{2}{c}{\bf 5 items}&\multicolumn{2}{c}{\bf 7 items}\\
\cmidrule(lr){3-4}
\cmidrule(lr){5-6}
\cmidrule(lr){7-8}
&&ACC&Avg ACC&ACC&Avg ACC&ACC&Avg ACC\\
\midrule
\multirow{3}{*}{\rotatebox[origin=c]{90}{\bf Base}}&1 condition&79.5&80.0&74.3&76.1&69.9&72.4\\
&2 conditions&51.4&55.7&33.2&44.9&21.0&38.9\\
&3 conditions&28.9&39.0&14.0&33.5&7.5&30.6\\
\midrule
\multirow{3}{*}{\rotatebox[origin=c]{90}{\bf Ours}}&1 condition&79.0&79.6&73.5&75.4&70.3&72.7\\
&2 conditions&55.7&61.8&41.0&53.8&31.1&49.7\\
&3 conditions&35.4&44.8&17.2&37.9&11.6&33.2\\
\bottomrule
\end{tabular}
\caption{o1-mini and o1-mini with EXSIR (ours) performances for paragraph-level setting in MCRank.}
\label{tab:o1-mini-par}
\end{table*}
\begin{table*}
\small
\centering
\begin{tabular}{llrrrrrr}
\toprule 
&\multirow{2}{*}{\bf Models}&\multicolumn{2}{c}{\bf 3 items}&\multicolumn{2}{c}{\bf 5 items}&\multicolumn{2}{c}{\bf 7 items}\\
\cmidrule(lr){3-4}
\cmidrule(lr){5-6}
\cmidrule(lr){7-8}
&&ACC&Avg ACC&ACC&Avg ACC&ACC&Avg ACC\\
\midrule
\multirow{6}{*}{\rotatebox[origin=c]{90}{\bf GPT-4}}&Positional&38.5&38.5&44.5&44.5&39.5&39.5\\
&Locational&83.2&83.2&49.2&49.2&81.4&81.4\\
&Temporal&72.0&75.7&53.5&66.1&46.5&57.1\\
&Trait-based&88.0&92.0&76.0&90.1&67.5&89.5\\
&Reason-based&91.5&92.5&88.5&88.5&86.5&86.5\\
&All&74.4&76.2&63.5&69.3&61.5&69.1\\
\midrule
\multirow{6}{*}{\rotatebox[origin=c]{90}{\bf ChatGPT}}&Positional&39.5&39.5&44.5&44.5&39.0&39.0\\
&Locational&74.0&74.0&51.7&51.7&69.5&69.5\\
&Temporal&63.5&67.0&35.0&47.0&27.5&36.6\\
&Trait-based&36.5&53.0&20.0&51.3&3.0&35.6\\
&Reason-based&84.5&85.3&76.5&76.5&76.5&76.5\\
&All&59.2&63.4&45.0&54.4&38.8&48.5\\
\midrule
\multirow{6}{*}{\rotatebox[origin=c]{90}{\bf Llama3.1}}&Positional&42.0 & 42.0 &47.0 & 47.0 &46.0 & 46.0\\
&Locational&87.2 & 87.2 &48.3 & 48.3 &83.0 & 83.0\\
&Temporal&66.5 & 69.6 &50.5 & 60.3 &45.0 & 55.2\\
&Trait-based&78.0 & 85.0 &40.0 & 72.7 &38.5 & 77.9\\
&Reason-based&94.0 & 94.8 &94.0 & 94.0 &94.0 & 94.0\\
&All&73.1 & 75.4 &57.7 & 69.3&56.6 & 65.9\\
\midrule
\multirow{6}{*}{\rotatebox[origin=c]{90}{\bf Mistral}}&Positional&31.0&31.0&18.0&18.0&18.5&18.5\\
&Locational&45.1&45.1&54.2&54.2&37.3&37.3\\
&Temporal&41.5&45.3&29.5&34.8&23.5&29.3\\
&Trait-based&53.0&68.9&20.5&55.0&5.0&38.0\\
&Reason-based&53.5&57.2&46.5&46.5&47.5&47.5\\
&All&44.8&49.6&31.9&40.6&24.6&33.6\\
\midrule
\multirow{6}{*}{\rotatebox[origin=c]{90}{\bf GPT-4-Ours}}&Positional&38.7&38.7&44.7&44.7&39.7&39.7\\
&Locational&83.0&83.0&48.9&48.9&81.2&81.2\\
&Temporal&72.2&75.6&53.4&65.9&46.3&57.0\\
&Trait-based&87.6&91.7&76.1&90.0&67.3&89.2\\
&Reason-based&91.3&92.3&88.4&88.4&86.3&86.3\\
&All&74.2&76.0&63.3&69.2&61.4&68.9\\
\midrule
\multirow{6}{*}{\rotatebox[origin=c]{90}{\bf ChatGPT-Ours}}&Positional&38.9&38.9&43.8&43.8&39.3&39.3\\
&Locational&73.7&73.7&51.3&51.3&69.8&69.8\\
&Temporal&63.0&66.6&34.2&46.3&27.3&36.3\\
&Trait-based&36.3&52.7&19.4&51.0&2.7&35.1\\
&Reason-based&84.3&85.1&76.2&76.2&76.1&76.1\\
&All&58.8&63.0&44.6&54.3&38.5&48.2\\
\midrule
\multirow{6}{*}{\rotatebox[origin=c]{90}{\bf Llama3.1-Ours}}&Positional&41.0&41.0&46.1&46.1&46.5&46.5\\
&Locational&85.7&85.7&49.5&49.3&84.4&84.4\\
&Temporal&67.8&71.3&52.2&61.8&43.5&53.4\\
&Trait-based&77.3&84.0&39.2&71.9&36.7&76.1\\
&Reason-based&93.1&93.5&95.0&95.0&92.6&92.6\\
&All&72.6&74.8&58.1&69.7&56.0&65.2\\
\midrule
\multirow{6}{*}{\rotatebox[origin=c]{90}{\bf Mistral-Ours}}&Positional&29.5&29.5&16.8&16.8&17.8&17.8\\
&Locational&44.6&44.6&53.3&53.3&36.4&36.4\\
&Temporal&42.5&46.8&28.5&33.0&22.5&27.8\\
&Trait-based&51.3&66.6&21.8&56.7&5.6&39.4\\
&Reason-based&52.5&55.8&45.4&45.4&46.5&46.5\\
&All&44.0&48.7&31.4&40.0&24.1&33.1\\
\bottomrule
\end{tabular}
\caption{Detailed breakdown of models performance for token-level items and 1 condition.}
\label{tab:token-1}
\end{table*}
\begin{prompt}[title={\footnotesize\texttt{Token-level CoT Ranking Prompt}}, label=prompt:token-cot]
Given following conditions: ``[string of conditions]'', sort the list of items ``[string of items]'' from left to right. Do not provide any explanation. \\
To sort the items, first extract the conditions, then sort the conditions based on their priority. Finally, apply the sorted conditions on the list of items iteratively updating their order in each iteration. Only report the final sorted list of items.
\end{prompt}
\begin{table*}
\small
\centering
\begin{tabular}{llrrrrrr}
\toprule 
&\multirow{2}{*}{\bf Models}&\multicolumn{2}{c}{\bf 3 items}&\multicolumn{2}{c}{\bf 5 items}&\multicolumn{2}{c}{\bf 7 items}\\
\cmidrule(lr){3-4}
\cmidrule(lr){5-6}
\cmidrule(lr){7-8}
&&ACC&Avg ACC&ACC&Avg ACC&ACC&Avg ACC\\
\midrule
\multirow{6}{*}{\rotatebox[origin=c]{90}{\bf GPT-4}}&Positional&30.6&33.9&20.0&34.1&6.7&25.2\\
&Locational&36.3&41.7&40.2&57.1&39.2&60.2\\
&Temporal&28.5&39.8&11.5&25.9&10.5&23.6\\
&Trait-based&19.0&28.2&17.5&35.1&3.5&24.1\\
&Reason-based&40.5&48.0&27.8&46.0&15.4&44.6\\
&All&30.8&38.2&21.5&37.7&10.5&30.0\\
\midrule
\multirow{6}{*}{\rotatebox[origin=c]{90}{\bf ChatGPT}}&Positional&25.9&31.6&13.0&27.0&7.3&22.4\\
&Locational&31.6&34.1&16.9&29.0&19.6&41.7\\
&Temporal&32.5&38.9&20.0&36.1&18.5&32.2\\
&Trait-based&19.0&30.3&7.0&27.1&0.5&24.7\\
&Reason-based&35.5&39.8&24.7&37.9&16.2&36.0\\
&All&28.8&35.0&15.9&31.5&10.8&29.2\\
\midrule
\multirow{6}{*}{\rotatebox[origin=c]{90}{\bf Llama3.1}}&Positional&39.4 & 35.4 &11.3 & 30.5 &2.7 & 22.5\\
&Locational&44.3 & 51.9 &12.5 & 38.0 &11.7 & 34.3\\
&Temporal&38.0 & 36.1 &11.0 & 23.7 &10.5 & 21.2\\
&Trait-based&29.0 & 42.6 &15.0 & 42.4 &4.5 & 31.5\\
&Reason-based&42.5 & 52.9 &19.7 & 41.2 &5.6 & 37.0\\
&All&34.3 & 43.5 &13.8 & 34.7 &6.3 & 27.7\\
\midrule
\multirow{6}{*}{\rotatebox[origin=c]{90}{\bf Mistral}}&Positional&25.9&35.2&9.7&25.8&6.7&18.8\\
&Locational&32.9&43.2&19.6&36.1&17.6&33.0\\
&Temporal&28.5&39.3&11.5&25.3&7.0&18.7\\
&Trait-based&23.5&40.7&4.0&32.4&2.5&23.8\\
&Reason-based&32.0&42.0&10.5&27.3&8.1&26.5\\
&All&28.4&40.0&10.2&28.8&6.6&22.3\\
\midrule
\multirow{6}{*}{\rotatebox[origin=c]{90}{\bf GPT-4-Ours}}&Positional&32.5&35.8&27.0&39.1&20.7&36.7\\
&Locational&43.9&49.1&50.0&60.2&35.3&55.7\\
&Temporal&25.5&37.0&17.5&34.4&10.0&27.1\\
&Trait-based&61.0&67.9&43.0&60.7&17.5&52.1\\
&Reason-based&46.5&52.6&37.3&52.1&22.0&51.0\\
&All&41.8&48.5&33.4&48.1&18.2&41.9\\
\midrule
\multirow{6}{*}{\rotatebox[origin=c]{90}{\bf ChatGPT-Ours}}&Positional&39.0&43.6&27.0&43.7&11.7&33.4\\
&Locational&42.7&50.3&23.2&43.2&17.6&34.9\\
&Temporal&42.5&52.0&19.5&36.9&18.5&33.4\\
&Trait-based&29.5&45.6&5.5&30.6&1.0&28.5\\
&Reason-based&40.5&48.7&25.3&40.4&13.8&36.9\\
&All&38.7&48.0&19.3&38.1&11.9&32.4\\
\midrule
\multirow{6}{*}{\rotatebox[origin=c]{90}{\bf Llama3.1-Ours}}&Positional&40.1 & 47.5 &25.9 & 46.3 &20.1 & 38.6\\
&Locational&46.7 & 53.4 &34.8 & 50.4 &13.7 & 41.8\\
&Temporal&20.0 & 30.0 &11.0 & 27.8 &7.5 & 22.5\\
&Trait-based&43.0 & 54.8 &19.5 & 44.5 &12.5 & 41.4\\
&Reason-based&47.0 & 55.4 &22.8 & 45.4 &13.8 & 47.8\\
&All&39.1 & 48.1 &21.5 & 41.9 &13.2 & 36.8\\
\midrule
\multirow{6}{*}{\rotatebox[origin=c]{90}{\bf Mistral-Ours}}&Positional&35.5&44.3&18.9&31.6&14.5&26.9\\
&Locational&25.1&37.5&20.5&34.1&19.6&29.1\\
&Temporal&25.5&41.0&17.0&32.0&18.0&28.0\\
&Trait-based&35.0&52.0&8.5&32.1&2.0&26.0\\
&Reason-based&33.5&44.1&18.5&36.5&10.6&28.0\\
&All&31.3&43.9&16.5&33.1&12.1&27.7\\
\bottomrule
\end{tabular}
\caption{Detailed breakdown of models performance for token-level items and 2 conditions.}
\label{tab:token-2}
\end{table*}
\begin{table*}
\small
\centering
\begin{tabular}{llrrrrrr}
\toprule 
&\multirow{2}{*}{\bf Models}&\multicolumn{2}{c}{\bf 3 items}&\multicolumn{2}{c}{\bf 5 items}&\multicolumn{2}{c}{\bf 7 items}\\
\cmidrule(lr){3-4}
\cmidrule(lr){5-6}
\cmidrule(lr){7-8}
&&ACC&Avg ACC&ACC&Avg ACC&ACC&Avg ACC\\
\midrule
\multirow{6}{*}{\rotatebox[origin=c]{90}{\bf GPT-4}}&Positional&9.0&26.5&1.8&26.1&1.3&22.6\\
&Locational&28.1&38.6&12.9&38.5&5.8&33.7\\
&Temporal&29.0&38.6&2.0&26.5&0.0&22.9\\
&Trait-based&31.0&40.9&11.5&34.8&3.0&31.3\\
&Reason-based&28.0&38.3&14.1&36.6&8.5&34.3\\
&All&25.1&37.6&7.6&31.6&2.3&27.1\\
\midrule
\multirow{6}{*}{\rotatebox[origin=c]{90}{\bf ChatGPT}}&Positional&12.2&26.8&4.2&24.5&0.0&24.4\\
&Locational&17.0&26.3&14.8&38.0&3.8&30.8\\
&Temporal&18.0&30.7&4.0&27.9&0.0&23.0\\
&Trait-based&19.5&30.2&3.0&25.4&0.0&22.6\\
&Reason-based&21.5&29.1&9.4&31.1&4.2&28.6\\
&All&17.7&28.7&6.0&28.1&0.6&24.3\\
\midrule
\multirow{6}{*}{\rotatebox[origin=c]{90}{\bf Llama3.1}}&Positional&10.0 & 31.0 &3.6 & 29.2 &0.0 & 26.2\\
&Locational&25.7 & 34.8 &5.5 & 31.4 &0.0 & 21.4\\
&Temporal&24.5 & 37.8 &2.5 & 19.3 &0.5 & 17.4\\
&Trait-based&23.5 & 38.3 &8.5 & 30.3 &4.5 & 29.5\\
&Reason-based&28.0 & 38.4 &10.3 & 34.1 &2.1 & 31.4\\
&All&22.3 & 36.1 &5.7 & 27.9 &1.6 & 24.5\\
\midrule
\multirow{6}{*}{\rotatebox[origin=c]{90}{\bf Mistral}}&Positional&13.8&28.0&1.2&18.0&1.3&12.9\\
&Locational&16.7&31.9&4.6&24.8&1.9&16.2\\
&Temporal&16.0&32.8&2.5&23.9&4.0&18.3\\
&Trait-based&14.0&32.9&2.5&25.2&1.5&22.6\\
&Reason-based&20.5&35.3&1.9&21.1&0.0&17.2\\
&All&16.2&32.2&2.4&22.7&2.1&18.1\\
\midrule
\multirow{6}{*}{\rotatebox[origin=c]{90}{\bf GPT-4-Ours}}&Positional&8.5&24.5&3.6&19.5&0.0&17.0\\
&Locational&35.6&45.6&33.3&51.3&17.3&40.7\\
&Temporal&33.5&44.6&8.5&35.0&2.5&29.2\\
&Trait-based&48.0&52.9&29.5&48.4&17.0&46.1\\
&Reason-based&43.0&50.7&24.5&43.9&23.4&52.3\\
&All&34.0&43.9&18.5&38.7&9.0&34.0\\
\midrule
\multirow{6}{*}{\rotatebox[origin=c]{90}{\bf ChatGPT-Ours}}&Positional&4.8&21.9&1.8&20.1&1.3&19.2\\
&Locational&22.2&36.8&12.0&39.4&3.8&28.6\\
&Temporal&31.5&44.6&3.5&27.3&0.5&21.6\\
&Trait-based&31.5&43.7&5.0&31.5&0.5&28.1\\
&Reason-based&26.5&40.8&6.6&30.9&6.3&29.5\\
&All&23.7&38.1&5.2&28.4&1.4&23.3\\
\midrule
\multirow{6}{*}{\rotatebox[origin=c]{90}{\bf Llama3.1-Ours}}&Positional&5.2 & 25.5 &0.0 & 20.8 &0.0 & 18.1\\
&Locational&33.3 & 44.0 &19.4 & 40.1 &3.8 & 28.8\\
&Temporal&30.0 & 42.9 &7.5 & 31.8 & 2.0 & 25.0\\
&Trait-based&44.0 & 52.3 &22.0 & 44.6 &18.0 & 46.2\\
&Reason-based&36.0 & 43.5 &19.8 & 36.0 &10.6 & 40.1\\
&All&29.8 & 41.8 &12.9 & 34.5 &7.1 & 31.3\\
\midrule
\multirow{6}{*}{\rotatebox[origin=c]{90}{\bf Mistral-Ours}}&Positional&15.3&33.1&5.5&22.7&5.2&16.5\\
&Locational&17.0&35.2&9.3&21.7&9.6&25.9\\
&Temporal&17.0&34.1&6.5&23.6&5.0&15.2\\
&Trait-based&15.5&33.6&3.0&20.2&1.0&17.8\\
&Reason-based&11.5&28.4&7.5&21.6&4.3&17.2\\
&All&15.2&32.8&5.8&21.8&4.5&17.3\\
\bottomrule
\end{tabular}
\caption{Detailed breakdown of models performance for token-level items and 3 conditions.}
\label{tab:token-3}
\end{table*}

\begin{table*}
\small
\centering
\begin{tabular}{llrrrrrr}
\toprule 
&\multirow{2}{*}{\bf Models}&\multicolumn{2}{c}{\bf 3 items}&\multicolumn{2}{c}{\bf 5 items}&\multicolumn{2}{c}{\bf 7 items}\\
\cmidrule(lr){3-4}
\cmidrule(lr){5-6}
\cmidrule(lr){7-8}
&&ACC&Avg ACC&ACC&Avg ACC&ACC&Avg ACC\\
\midrule
\multirow{6}{*}{\rotatebox[origin=c]{90}{\bf GPT-4}}&Positional&44.0&44.0&46.0&46.0&43.5&43.5\\
&Locational&96.5&96.5&92.5&92.5&95.5&95.5\\
&Temporal&58.0&64.8&49.0&54.2&51.5&55.9\\
&Trait-based&85.5&87.7&82.0&82.5&77.0&77.8\\
&Reason-based&28.0&36.0&22.0&27.9&14.5&17.4\\
&All&62.4&65.8&58.3&60.6&56.4&58.0\\
\midrule
\multirow{6}{*}{\rotatebox[origin=c]{90}{\bf ChatGPT}}&Positional&25.5&25.5&26.5 & 26.5&26.5 & 26.5\\
&Locational&59.5&59.5&49.5 & 49.5&43.5 & 43.5\\
&Temporal&38.0&42.5&17.0 & 25.9&12.0 & 16.4\\
&Trait-based&60.0&61.2&39.5 & 40.4&45.5 & 46.6\\
&Reason-based&32.5&38.2&13.5 & 15.7&12.0 & 14.7\\
&All&43.1&45.4&29.2&31.6&27.9&29.6\\
\midrule
\multirow{6}{*}{\rotatebox[origin=c]{90}{\bf Llama3.1}}&Positional&40.5 & 40.5 &44.0 & 44.0 & 42.0 & 42.0\\
&Locational&96.5 & 96.5 &97.0 & 97.0 &95.0 & 95.0\\
&Temporal&62.0 & 65.6 & 46.0 & 54.0 &29.5 & 40.9\\
&Trait-based&85.0 & 86.3 &79.0 & 81.1 &73.0 & 74.7\\
&Reason-based&37.5 & 42.4 &19.0 & 25.4 &15.0 & 20.2\\
&All&64.3 & 66.3 &57.0 & 60.3 &50.9 & 54.5\\
\midrule
\multirow{6}{*}{\rotatebox[origin=c]{90}{\bf Mistral}}&Positional&40.5&40.5&29.5&29.5&33.5&33.5\\
&Locational&41.5&41.5&41.0&41.0&33.0&33.0\\
&Temporal&26.0&33.0&13.5&21.5&9.5&14.3\\
&Trait-based&48.5&49.8&42.5&44.7&35.0&35.6\\
&Reason-based&26.0&32.2&9.0&13.6&11.5&14.6\\
&All&36.5&39.4&27.1&30.1&24.5&26.2\\
\midrule
\multirow{6}{*}{\rotatebox[origin=c]{90}{\bf GPT-4-Ours}}&Positional&43.3&43.3&46.4&46.4&43.6&43.6\\
&Locational&97.1&97.1&93.1&93.1&96.2&96.2\\
&Temporal&57.2&63.9&48.4&53.7&50.9&55.0\\
&Trait-based&84.6&86.6&81.3&81.6&78.0&76.8\\
&Reason-based&27.2&34.9&21.5&27.0&13.8&16.4\\
&All&61.9&65.2&58.1&60.3&56.1&57.6\\
\midrule
\multirow{6}{*}{\rotatebox[origin=c]{90}{\bf ChatGPT-Ours}}&Positional&24.5&24.5&26.2 & 26.2&25.8 & 25.8\\
&Locational&60.7&60.7&48.6 & 48.6&44.1 & 44.1\\
&Temporal&39.9&44.5&18.0 & 27.4&13.1 & 17.6\\
&Trait-based&58.3&59.2&38.3 & 38.9&44.5 & 45.1\\
&Reason-based&31.0&36.3&12.5 & 14.7&11.4 & 13.5\\
&All&42.9&45.1&29.1&31.5&27.8&29.3\\
\midrule
\multirow{6}{*}{\rotatebox[origin=c]{90}{\bf Llama3.1-Ours}}&Positional&40.0 & 40.0 &43.0 & 43.0 &42.0 & 42.0\\
&Locational&97.5 & 97.5 &97.5 & 97.5 &98.0 & 98.0\\
&Temporal& 60.0 & 63.3 &47.0 & 52.7 &32.0 & 40.4\\
&Trait-based&87.5 & 88.5 &80.5 & 82.3 &70.5 & 72.2\\
&Reason-based& 34.5 & 40.0 &19.5 & 28.3 &20.0 & 25.4\\
&All&63.9 & 65.8 &57.5 & 60.7 &52.5 & 55.6\\
\midrule
\multirow{6}{*}{\rotatebox[origin=c]{90}{\bf Mistral-Ours}}&Positional&39.3&39.3&30.5&30.5&34.2&34.2\\
&Locational&42.2&42.2&41.5&41.5&33.0&33.0\\
&Temporal&25.0&31.8&12.5&20.1&8.5&12.8\\
&Trait-based&47.5&47.8&42.5&43.4&34.5&34.5\\
&Reason-based&24.7&29.8&9.0&12.4&10.8&13.8\\
&All&35.7&38.6&26.8&29.6&24.2&25.7\\
\bottomrule
\end{tabular}
\caption{Detailed breakdown of models performance for paragraph-level items and 1 condition.}
\label{tab:para-1}
\end{table*}

\begin{table*}
\small
\centering
\begin{tabular}{llrrrrrr}
\toprule 
&\multirow{2}{*}{\bf Models}&\multicolumn{2}{c}{\bf 3 items}&\multicolumn{2}{c}{\bf 5 items}&\multicolumn{2}{c}{\bf 7 items}\\
\cmidrule(lr){3-4}
\cmidrule(lr){5-6}
\cmidrule(lr){7-8}
&&ACC&Avg ACC&ACC&Avg ACC&ACC&Avg ACC\\
\midrule
\multirow{6}{*}{\rotatebox[origin=c]{90}{\bf GPT-4}}&Positional&27.5&36.1&14.0&32.0&6.5&30.0\\
&Locational&42.0&51.9&21.0&39.3&18.0&37.6\\
&Temporal&26.5&39.1&11.5&32.4&13.0&29.9\\
&Trait-based&40.0&49.6&27.0&44.6&14.5&30.7\\
&Reason-based&24.5&38.2&12.0&30.8&7.5&24.2\\
&All&32.1&43.0&17.1&35.8&11.9&30.5\\
\midrule
\multirow{6}{*}{\rotatebox[origin=c]{90}{\bf ChatGPT}}&Positional&19.0 & 25.7&8.5 & 20.9&6.0 &18.7\\
&Locational&14.0 & 24.2&7.5 & 23.5&6.0 & 19.2\\
&Temporal&15.0 & 26.3&6.0 & 19.0&1.0 & 11.9\\
&Trait-based&23.0 & 32.5&10.5 & 27.7&5.0 & 19.4\\
&Reason-based&14.0 & 27.1&2.5 & 16.4&1.5 & 10.9\\
&All&17.0&27.1&7.0&21.5&3.9& 16.0\\
\midrule
\multirow{6}{*}{\rotatebox[origin=c]{90}{\bf Llama3.1}}&Positional&25.0 & 36.2 &12.5 & 30.2 &6.0 & 24.5\\
&Locational&41.0 & 52.5 &25.5 & 41.9 &18.5 & 36.1\\
&Temporal&26.5 & 42.2 &11.0 & 30.2 &10.0 & 26.4\\
&Trait-based&34.5 & 45.8 &24.0 & 39.8 &13.0 & 27.8\\
&Reason-based& 24.0 & 36.4 &9.5 & 26.4 &4.0 & 21.1\\
&All& 30.2 & 42.6 &16.5 & 33.7 &10.3 & 27.2\\
\midrule
\multirow{6}{*}{\rotatebox[origin=c]{90}{\bf Mistral}}&Positional&25.0&36.8&11.5&26.5&3.5&19.2\\
&Locational&17.0&31.6&5.0&23.3&6.0&18.4\\
&Temporal&9.5&26.7&6.5&23.0&4.0&19.5\\
&Trait-based&18.0&34.3&7.5&26.4&7.0&20.8\\
&Reason-based&8.5&26.0&4.5&19.6&3.0&16.5\\
&All&15.6&31.1&7.0&23.8&4.7&18.9\\
\midrule
\multirow{6}{*}{\rotatebox[origin=c]{90}{\bf GPT-4-Ours}}&Positional&32.0&43.9&11.5&34.9&7.0&31.5\\
&Locational&62.5&68.8&33.0&53.7&27.5&48.7\\
&Temporal&44.0&55.2&17.5&37.8&10.0&33.3\\
&Trait-based&53.0&59.6&28.0&50.6&22.5&42.8\\
&Reason-based&30.5&43.2&11.5&28.6&7.5&24.0\\
&All&44.4&54.1&20.3&41.1&14.9&36.1\\
\midrule
\multirow{6}{*}{\rotatebox[origin=c]{90}{\bf ChatGPT-Ours}}&Positional&20.0 & 30.1&9.5 & 22.4&5.0 & 17.7\\
&Locational&22.0 & 34.3&11.5 & 31.4&6.5 & 23.0\\
&Temporal&22.5 & 38.9&4.5 & 23.4&3.5 & 19.0\\
&Trait-based&27.0 & 42.6&12.0 & 30.4&9.0 & 24.6\\
&Reason-based&18.0 & 34.3&7.0 & 24.8&4.0 & 17.5\\
&All&21.9 & 36.1&8.9&26.5&5.6 & 20.3\\
\midrule
\multirow{6}{*}{\rotatebox[origin=c]{90}{\bf Llama3.1-Ours}}&Positional&34.0 & 44.4 &16.0 & 33.0 &6.0 & 27.3\\
&Locational&50.5 & 60.3 &22.5 & 46.0 &21.0 & 45.0\\
&Temporal&34.5 & 46.2 &11.5 & 31.0 &3.5 & 19.0\\
&Trait-based&43.0 & 53.7 &26.0 & 44.1 &15.5 & 29.6\\
&Reason-based&23.0 & 36.3 &8.0 & 24.9 &4.0 & 16.5\\
&All&37.0 & 48.2 &16.8 & 35.8 &10.0 & 27.5\\
\midrule
\multirow{6}{*}{\rotatebox[origin=c]{90}{\bf Mistral-Ours}}&Positional&18.5&31.8&5.0&23.5&3.0&21.6\\
&Locational&22.0&33.0&5.0&16.8&4.5&15.5\\
&Temporal&20.0&34.1&5.5&22.4&4.0&16.4\\
&Trait-based&17.5&34.3&7.0&24.5&5.5&20.5\\
&Reason-based&11.0&25.8&5.5&20.8&1.0&14.7\\
&All&17.8&31.8&5.6&21.6&3.6&17.8\\
\bottomrule
\end{tabular}
\caption{Detailed breakdown of models performance for paragraph-level items and 2 conditions.}
\label{tab:para-2}
\end{table*}

\begin{table*}
\small
\centering
\begin{tabular}{llrrrrrr}
\toprule 
&\multirow{2}{*}{\bf Models}&\multicolumn{2}{c}{\bf 3 items}&\multicolumn{2}{c}{\bf 5 items}&\multicolumn{2}{c}{\bf 7 items}\\
\cmidrule(lr){3-4}
\cmidrule(lr){5-6}
\cmidrule(lr){7-8}
&&ACC&Avg ACC&ACC&Avg ACC&ACC&Avg ACC\\
\midrule
\multirow{6}{*}{\rotatebox[origin=c]{90}{\bf GPT-4}}&Positional&10.5&26.3&2.0&24.8&1.0&25.5\\
&Locational&30.5&35.9&12.0&34.2&1.0&24.3\\
&Temporal&24.5&39.2&7.0&31.7&0.0&20.3\\
&Trait-based&36.5&44.9&8.5&33.4&1.0&26.1\\
&Reason-based&25.5&41.8&2.5&27.1&0.0&20.5\\
&All&25.5&37.6&6.4&30.2&0.6&23.3\\
\midrule
\multirow{6}{*}{\rotatebox[origin=c]{90}{\bf ChatGPT}}&Positional&9.5 & 17.3&1.5 & 14.5&0.0 & 10.7\\
&Locational&11.5 & 19.5&3.5 & 17.1&0.0 & 9.2\\
&Temporal&9.5 & 20.0 & 0.0 & 17.6&0.0 & 12.8\\
&Trait-based&19.0 & 30.2&1.5 & 20.1&0.0 & 16.5\\
&Reason-based&12.0 & 28.3&0.2 & 18.3&0.0 & 16.1\\
&All&12.3& 23.1&1.7 & 17.5&0.0 &13.1\\
\midrule
\multirow{6}{*}{\rotatebox[origin=c]{90}{\bf Llama3.1}}&Positional&7.0 & 27.6 &2.0 & 26.6 &1.0 & 22.0\\
&Locational&33.0 & 41.9 &9.5 & 35.6 & 0.0 & 23.7\\
&Temporal&26.0 & 42.4 &2.5 & 27.4 &0.5 & 18.5\\
&Trait-based&31.5 & 42.8 &8.0 & 31.4 &0.5 & 22.4\\
&Reason-based&29.5 & 43.3 &2.5 & 24.7 &0.0 & 18.7\\
&All&25.4 & 39.6 &4.9 & 29.1 &0.4 & 21.1\\
\midrule
\multirow{6}{*}{\rotatebox[origin=c]{90}{\bf Mistral}}&Positional&13.0&33.3&1.0&25.0&0.5&17.5\\
&Locational&6.0&26.3&0.5&18.8&0.0&15.0\\
&Temporal&6.5&25.3&0.5&17.3&0.0&14.6\\
&Trait-based&11.0&27.5&1.0&21.8&0.0&15.4\\
&Reason-based&7.0&26.0&0.5&20.9&0.0&13.6\\
&All&8.7&27.7&0.7&20.8&0.1&15.2\\
\midrule
\multirow{6}{*}{\rotatebox[origin=c]{90}{\bf GPT-4-Ours}}&Positional&10.5&29.3&0.5&22.2&0.5&19.5\\
&Locational&38.5&45.3&13.0&40.9&3.5&31.8\\
&Temporal&36.5&47.0&10.0&30.4&3.0&24.3\\
&Trait-based&47.5&53.7&12.0&35.7&2.0&28.1\\
&Reason-based&32.5&47.5&4.5&29.6&0.0&19.9\\
&All&33.1&44.6&8.1&31.8&1.8&24.7\\
\midrule
\multirow{6}{*}{\rotatebox[origin=c]{90}{\bf ChatGPT-Ours}}&Positional&22.5 & 38.0&2.5 & 22.5&0.0 & 13.1\\
&Locational&14.0 & 26.3&3.5 & 21.1&2.0 & 18.8\\
&Temporal&18.5 & 33.8&2.5 & 22.3&0.0 & 16.4\\
&Trait-based&16.0 & 34.5&2.0 & 23.3&0.5 & 18.7\\
&Reason-based&14.0 & 31.4&2.0 & 22.1&0.0 & 18.8\\
&All&17.0& 32.8&2.5 & 22.2&0.5&17.2\\
\midrule
\multirow{6}{*}{\rotatebox[origin=c]{90}{\bf Llama3.1-Ours}}&Positional&29.5 & 45.8 &2.5 & 27.1 &0.0 & 17.2\\
&Locational&25.5 & 42.1 &10.0 & 31.5 &0.5 & 18.7\\
&Temporal&25.5 & 44.4 &1.5 & 24.4 &0.0 & 16.7\\
&Trait-based&26.5 & 43.1 &9.0 & 36.2 &0.0 & 22.5\\
&Reason-based&22.5 & 33.5 &2.5 & 23.7 &0.5 & 16.5\\
&All&25.9 & 41.8 &5.1 & 28.5 &0.2 & 18.3\\
\midrule
\multirow{6}{*}{\rotatebox[origin=c]{90}{\bf Mistral-Ours}}&Positional&15.5&34.7&1.0&21.5&0.0&17.5\\
&Locational&14.5&29.5&1.0&17.3&0.0&13.0\\
&Temporal&15.0&29.2&1.0&20.3&0.0&15.8\\
&Trait-based&13.0&29.5&0.5&18.7&0.0&14.6\\
&Reason-based&11.0&25.8&2.5&23.9&0.0&13.9\\
&All&13.8&29.7&1.2&20.3&0.0&15.0\\
\bottomrule
\end{tabular}
\caption{Detailed breakdown of models performance for paragraph-level items and 3 conditions.}
\label{tab:para-3}
\end{table*}


\end{document}